\documentclass[runningheads]{llncs}
\usepackage[T1]{fontenc}
\usepackage{graphicx,verbatim}
\usepackage{llncs-custom}
\newcommand{\cls}{\texttt{[CLS]}}
\newcommand{\hres}[2]{#1 {\scriptsize $\pm$ #2}}
\begin{document}
\title{Pulmonary Embolism Risk Stratification from CTPA and Medical Records: Vascular Graphs Are Not All You Need}
\titlerunning{Pulmonary Embolism Risk Stratification from CTPA and Medical Records}
%

\author{Nathan~Painchaud\inst{1}\textsuperscript{(\Letter)}
\and
Tristan~Habémont\inst{1} \and
Morgane~des~Ligneris\inst{1}
\and
Allan~Serva\inst{2,3} \and
Pierre~Croisille\inst{1,4}
\and
Laurent~Bertoletti\inst{2}
\and
Thomas~Lampert\inst{5}
\and
Johannes~F.~Lutzeyer\inst{6}
\and
Odyssée~Merveille\inst{1}
}

%
\authorrunning{N. Painchaud et al.}
%
\institute{INSA‐Lyon, Université Lyon 1, UJM-Saint Etienne, CNRS, Inserm, CREATIS UMR 5220, U1294, F‐69621, Lyon, France \and
Université Jean Monnet Saint-Étienne, CHU Saint-Étienne, Mines Saint-Étienne, INSERM, SAINBIOSE U1059, CIC 1408, Département de Médecine Vasculaire et Thérapeutique, F-CRIN INNOVTE network, all in F-42055, Saint-Étienne, France \and
Department of Pneumology, CHU Saint-Etienne, UJM Saint-Etienne, France \and
Department of Radiology, CHU Saint-Etienne, UJM Saint-Etienne, France \and
ICube, University of Strasbourg, France \and
LIX, CNRS, École Polytechnique, Institut Polytechnique de Paris, Palaiseau, France \\
\email{nathan.painchaud@insa-lyon.fr} }

\maketitle              
\begin{abstract}
Risk stratification for pulmonary embolism (PE) is critical for clinical decision-making. Stratification guidelines are based on patient medical records, parameters measured from computed tomography pulmonary angiography (CTPA), and blood tests. However, blood tests are often missing in routine practice. This work studies whether \sota\ models can accurately classify risk stratification from only medical records and biomarkers extracted from CTPA images. We benchmark different approaches to combine medical records and cardiac biomarkers with rich pulmonary vascular information; we add vascular biomarkers to tabular models and apply graph neural networks (GNNs) on the vascular tree's intrinsic graph representation. We use a private dataset (n=353) with uniquely complete data for PE risk stratification. Our results show that, among global features, medical records and cardiac biomarkers are the most significant predictors, while vascular biomarkers do not further improve stratification. Even more surprising, even GNNs on vascular graphs fail to outperform strong tabular baseline on global features. We consider hypotheses, on both models and data, that could explain this suboptimal performance. Our investigation suggests that, counter-intuitively, vascular graphs might hold no discriminative information for PE risk stratification. Code is available from \url{https://github.com/creatis-myriad/GENESIS}.


\end{abstract}
\section{Introduction}
\label{sec:introduction}
Risk stratification for pulmonary embolism (PE) determines how patients are treated, e.g., whether they receive intensive care, thrombolytic therapy, or anticoagulants. Current stratification guidelines rely on a combination of medical records, visual evaluation of Computed Tomography Pulmonary Angiography (CTPA), and blood tests to measure troponin and NT-proBNP~\cite{konstantinides_2019_2020}. CTPA is the standard diagnostic modality and is almost universally available, but blood tests are frequently not performed for the initial diagnosis~\cite{liu_joint_2021}. This gap between guidelines and clinical practice motivates research on maximising the diagnostic accuracy from CTPA and medical records only.

While CTPA is used to confirm PE diagnosis by the presence of thrombi, its current use for risk stratification is limited to measuring right ventricle (RV) heart dysfunction, rather than thrombus burden. Clinical scores were proposed to globally assess thrombus burden from CTPA, e.g., Qanadli~\cite{qanadli_new_2001} and Mastora~\cite{mastora_severity_2003} vascular obstruction scores, and were correlated to PE risk factors. However, they require to locate thrombi's positions in the pulmonary arterial tree, a labour too time-intensive to perform manually in routine clinical settings.

Deep neural networks have been successfully applied to automate part of this process by detecting~\cite{soffer_deep_2021} and segmenting~\cite{djahnine_detection_2024} PE in CTPA. They have also been used to segment the pulmonary arterial tree~\cite{chu_deep_2025,liu_custom_2025}. Recently, an automated pipeline was introduced to segment both the arterial tree and thrombi, and to extract a feature-rich patient-specific graph representation of the arterial tree~\cite{des-ligneris_patient-specific_2026}. Still, since thrombi and arteries are thin and sparse in CTPA, the spatial priors of deep vision models make them ill-suited for patient-level PE risk stratification.

By contrast, the new patient-specific vascular graphs and the rising popularity of graph neural networks (GNN) for medical applications~\cite{ahmedt-aristizabal_graph-based_2021} present the opportunity to use a framework intrinsically adapted to the application. However, global features, i.e., patient-level data and biomarkers, are important for PE risk stratification~\cite{konstantinides_2019_2020}, and comparatively little research has explored multimodal learning with GNNs, all applications considered~\cite{brasoveanu_extending_2023,sun_graph_2024,lutchyn_efficient_2025}. Furthermore, no specific GNN model dominates across tasks and datasets~\cite{liu_graph_2025}, and even graph foundation models (GFMs) for healthcare currently target specific tasks, mostly knowledge discovery and brain networks~\cite{khan_comprehensive_2025}.

In this context, specifying an optimal PE risk stratification model is not trivial. Using a private dataset with uniquely complete PE risk stratification data, we propose the first study of the usefulness of pulmonary vascular graphs for PE risk stratification, structured into three main contributions:
\begin{enumerate}
    \item We introduce a pipeline for automatic PE risk stratification from routine clinical data, i.e., structured medical records and CTPA images;
    \item We benchmark \sota\ tabular models and GNNs to compare two approaches for adding vascular information: global biomarkers vs.\ full graphs. We test several backbones, notably two new proposed methods to combine global features in GNNs;
    \item We empirically investigate hypotheses to explain counter-intuitive negative results regarding the relevance of vascular graphs for PE risk stratification.
\end{enumerate}

\section{Methods}
\label{sec:methods}
\begin{figure}
    \centering
    \includegraphics[width=\linewidth]{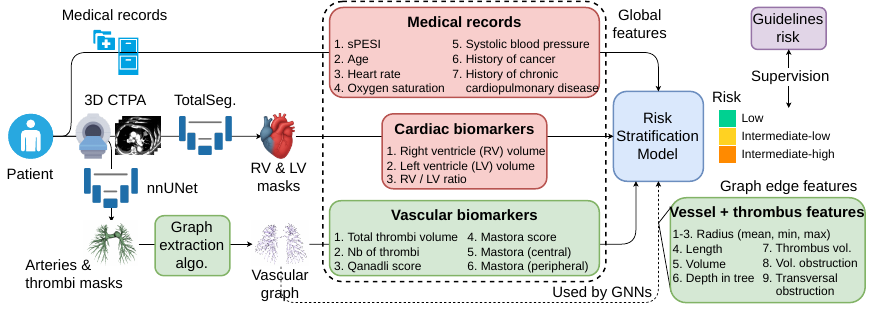}
    \caption{Proposed pipeline for PE risk stratification, including CTPA preprocessing.}
    \label{fig:persevere_pipeline}
\end{figure}

We propose a pipeline, detailed in \cref{fig:persevere_pipeline}, to categorize PE risk in three classes by combining routine clinical data, i.e., structured medical records and CTPA images. Clinical guidelines were used to determine reference PE risk, using blood tests results not provided to our pipeline.

We first preprocess high-dimensional 3D CTPA by segmenting structures of interest for PE, i.e., ventricles and pulmonary vasculature. The latter is then converted into a graph. Finally, cardiac and vascular biomarkers are computed from the ventricle masks and vascular graph, respectively (see \cref{sec:methods:preprocessing}).

\begin{figure}[t]
    \subfloat[Baseline tabular foundation model~\cite{hollmann_accurate_2025} on global features.\label{fig:fusion_methods:tabular}]{
        \includegraphics[width=0.43\linewidth]{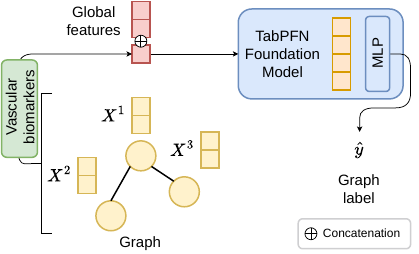}
    }
    \hspace{6pt}
    \subfloat[Existing methods of early fusion~\cite{lutchyn_efficient_2025} and \\ late fusion~\cite{brasoveanu_extending_2023,sun_graph_2024}.\label{fig:fusion_methods:early+late}]{
        \includegraphics[width=0.53\linewidth]{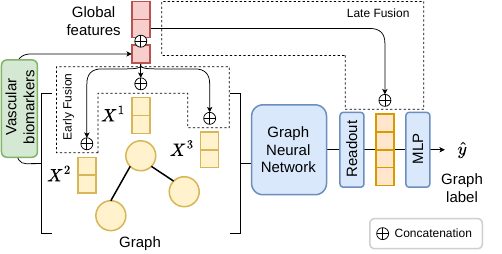}
    }
    \hfill
    \subfloat[Virtual node~\cite{pham_graph_2017} with proposed \\ initialization from global features.\label{fig:fusion_methods:virtual_node}]{
        \includegraphics[width=0.45\linewidth]{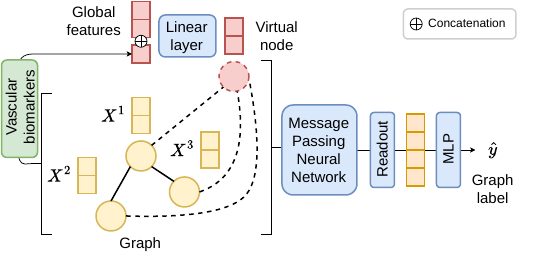}
    }
    \subfloat[Graph Transformer~\cite{rampasek_recipe_2022} with proposed \\ cross-attention on tokenized global features~\cite{gorishniy_revisiting_2021}.\label{fig:fusion_methods:gt_cross_attention}]{
        \includegraphics[width=0.54\linewidth]{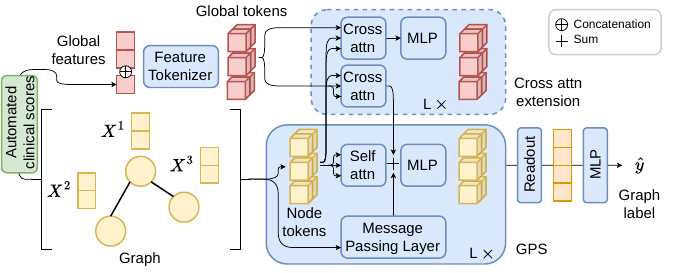}
    }
    \caption{Schemas summarizing benchmarked methods for combining global features and vascular graphs for patient risk classification. Some layers (e.g., residual connections, norm, dropout) are not shown for readability.}
    \label{fig:fusion_methods}
\end{figure}

The stratification model then combines structured medical records and extracted image features. Since medical records and CTPA are complementary, we focus on exploring two fusion strategies, which tend to perform better in this context \cite{guarrasi_systematic_2025}. We test \emph{tabular models} where image information is provided by concatenating cardiac and vascular biomarkers to medical records, as shown in \cref{fig:fusion_methods:tabular}, motivated by similar applications having shown tabular models to be strong baselines~\cite{stym-popper_dafted_2025}. We also test \emph{GNNs}, extended to support global features (see \cref{sec:methods:graphs}), to make use of the rich vascular graph representation.

\subsection{CTPA Images Preprocessing \& Vascular Graph Representation}
\label{sec:methods:preprocessing}
To segment the heart's left and right ventricles (LV \& RV), we use TotalSegmentator~\cite{wasserthal_totalsegmentator_2023}, a reference nnUNet~\cite{isensee_nnu-net_2021} pretrained to segment important anatomical structures in CT. From these masks, we compute 3 cardiac biomarkers related to RV function (middle branch in \cref{fig:persevere_pipeline}).

To extract the pulmonary arterial graph, we use the pipeline developed in~\cite{des-ligneris_patient-specific_2026}. An nnUNet was iteratively trained and its predictions semi-automatically checked to obtain anatomically and topologically consistent masks of arteries and thrombi. The arteries masks were converted to graphs by representing junctions or terminal vessels as nodes and vessels as edges, with automatic quality checks and corrections to ensure topological consistency (bottom left in \cref{fig:persevere_pipeline}). Each vessel, i.e., edge in the graph, is described using 6 features computed from the masks (bottom right in \cref{fig:persevere_pipeline}). Thrombi are described by concatenating 3 thrombus features, i.e., thrombus volume and obstructions, to the 6 vessel features. These features are equal to 0 in the absence of thrombi.

Finally, the vascular graph enables the automatic computation of 6 patient-level vascular biomarkers defined by clinicians to summarise thrombus burden (bottom centre in \cref{fig:persevere_pipeline}), previously too time-consuming to measure manually.

\subsection{Graph Neural Networks with Global Features}
\label{sec:methods:graphs}
Compared to the extensive research on multimodal fusion in computer vision, few works have studied how to combine global features with graphs. Previous works have proposed two fusion methods (see \cref{fig:fusion_methods:early+late}), namely \emph{early fusion} by concatenating global features on each node~\cite{lutchyn_efficient_2025} and \emph{late fusion} by concatenating global features to the graph embedding of a GNN encoder~\cite{brasoveanu_extending_2023,sun_graph_2024}. Considering computer vision literature has shown \emph{intermediate fusion} typically performs better than early or late fusion when data modalities are structured differently~\cite{guarrasi_systematic_2025}, we propose two new intermediate fusion methods designed for GNNs.

\smallskip
\noindent{\textbf{Virtual Node from Global Features. }}
Previous works have proposed to facilitate interactions between faraway nodes by adding virtual nodes (VN), i.e., nodes not originally in the graph, connected to all other nodes~\cite{pham_graph_2017}. VNs have primarily been used as they tend to improve performance on graph-level tasks. Prior works also suggested to use VNs to integrate global features and store a global state~\cite{pham_graph_2017}, but did not test this in practice. Thus, we perform the first empirical evaluation of VNs initialised from global features, as shown in \cref{fig:fusion_methods:virtual_node}.

\smallskip
\noindent{\textbf{Cross-attention in Graph Transformers. }}
Graph Transformers (GT) have become a \sota\ alternative to message passing neural networks (MPNN), demonstrating high performance on graph-level tasks~\cite{rampasek_recipe_2022}. Additionally, Transformers' ability to adapt to data with different structures and levels of details helped them succeed in multimodal applications~\cite{xu_multimodal_2023}. These successes motivated us to propose to use cross-attention in GTs to fuse global features.

We use the \sota\ GPS~\cite{rampasek_recipe_2022} as the backbone. It augments the graph by adding a node's positional encoding to its features. Node features are updated in parallel by an MPNN layer and self-attention between nodes treated as a sequence of tokens. On top of GPS, we propose to use a Feature Tokeniser~\cite{gorishniy_revisiting_2021} to map each global feature to a token, and to perform bidirectional cross-attention between global tokens and node tokens, as shown in \cref{fig:fusion_methods:gt_cross_attention}.

\section{Experiments \& Results}
\label{sec:experiment+results}

\subsection{Experimental Settings}
\label{sec:experiment+results:settings}

\noindent{\textbf{Dataset. }}
Data were collected from 353 patients diagnosed with PE by CTPA at the CHU Saint-Étienne emergency department between April 2014 and June 2020. The retrospective study was approved by the local ethics committee \linebreak (IRBN262017/CHUSTE). CTPA acquisitions details are provided in~\cite{des-ligneris_patient-specific_2026}. Compared to other PE datasets, additional information to stratify risk using ESC guidelines~\cite{konstantinides_2019_2020} was collected: troponin \& BNP measurements, sPESI score and its 6 factors~\cite{jimenez_simplification_2010}. Risk groups are: low (n=88), intermediate-low (n=138), and intermediate-high (n=127). To evaluate models, 10-fold cross-validation with an 80/10/10 split was used.

\smallskip
\noindent{\textbf{Tabular Models. }}
Since tree-based models have historically outperformed deep learning~\cite{gorishniy_revisiting_2021} we compare XGBoost~\cite{chen_xgboost_2016}, a reference machine learning method, against TabPFN~\cite{hollmann_accurate_2025}, a \sota\ Transformed-based foundation model.

\smallskip
\noindent{\textbf{Graph Neural Networks and Architecture Search. }}
All GNNs were implemented using PyTorch Geometric~\cite{fey_pyg_2025}. We conducted an exhaustive hyperparameter search across the options detailed in \cref{tab:hparams_search} using Optuna~\cite{akiba_optuna_2019} with the Tree-structured Parzen Estimator (TPE) and a budget of 250 cross-validation trials for each GNN backbone. For each backbone, we chose the configuration with the best validation average across folds.

Additionally, we ran an ablation study of the best GNN configurations' key components, not detailed because of space constraints. Results showed that using the line graphs, i.e., inverting edges and vertices, improves performance, explained by the graphs' features being exclusively on edges. They also confirmed that embedding global features using a Feature Tokenizer~\cite{gorishniy_revisiting_2021} (see \cref{fig:fusion_methods:gt_cross_attention}) outperforms both linear projection and TabPFN's internal embedding.

\begin{table}[tb]
\centering
\caption{Configuration space for automated hyperparameter tuning of GNNs.}
\label{tab:hparams_search}
\begin{NiceTabular}{l cc}
\toprule
\Block{2-1}{Hyperparameter} & \Block{1-2}{Tested configurations} \\
\cmidrule(lr){2-3}
& Message passing GNN (MPNN) & Graph Transformer (GT) \\
\midrule
Backbone & \makecell{ GCN~\cite{kipf_semi-supervised_2017}, GATv2~\cite{brody_how_2021}, GIN~\cite{xu_how_2018} \\ with/without Virtual Node (VN) } & GPS~\cite{rampasek_recipe_2022} \\
Attention heads & -- & 1, 2, 4 \\
\makecell[l]{Graph representation \\ (readout)} & \makecell{Global pooling: sum, mean, \\ virtual node} & \makecell{Global pooling: sum, mean \\ Tokens: \cls\ token} \\
\midrule
Layer depth & \Block{1-2}{ $2 \dots 10$ } \\
Hidden channels & \Block{1-2}{ 16, 32, 64 } \\
Normalization & \Block{1-2}{ BatchNorm, InstanceNorm, GraphNorm } \\
Dropout & \Block{1-2}{ $[0, 0.6]$ } \\
Prediction head layers & \Block{1-2}{ $1 \dots 5$ } \\
L2 regularization & \Block{1-2}{ $[10^{-5}, 10^{-3}$] } \\
\bottomrule
\end{NiceTabular}
\end{table}

\subsection{Key Results}
\label{sec:experiment+results:key}
In this section, we present key results on the relevance of pulmonary vascular graphs for PE risk stratification. We first study whether clinicians' vascular biomarkers improve tabular models. Then, we test if GNNs applied to the intrinsic vascular graph manage to use the richer data representation to improve upon the tabular baseline. Finally, we investigate whether data and models limitations explain the unexpected lack of improvement from vascular graphs.

Unless otherwise specified, scores in \cref{tab:biomarkers_ablation_study,tab:gnn_benchmark,tab:gnn_sanity_check} correspond to cross-validation test mean ± s.d. Results in \textbf{bold} are not statistically different from the best (p>0.05 using a one-sided Wilcoxon signed-rank test).

\smallskip
\noindent\textbf{Can tabular models learn from vascular biomarkers?}
Our first experiment studies the predictive power of global features. Cardiac biomarkers should be important, considering the link between RV dysfunction and PE risk~\cite{konstantinides_2019_2020}. Similarly, vascular biomarkers extracted from the graph should be predictive, given correlations between obstruction scores~\cite{mastora_severity_2003,qanadli_new_2001} and PE risk factors.

\begin{table}[tb]
\centering
\caption{Ablation study on the types of global features considered, using tabular models. Only the best model for each feature combination is presented.}
\label{tab:biomarkers_ablation_study}
\begin{NiceTabular}{c ccc c}
\toprule
Model & \Block{1-3}{Global features}&& & \Block{3-1}{F1 (\%) $\uparrow$} \\
\cmidrule(lr){1-1} \cmidrule(lr){2-4}
\makecell{Best \\ backbone} & \makecell{Medical \\ records} & \makecell{Cardiac \\ biomarkers} & \makecell{Vascular \\ biomarkers} & \\
\midrule
TabPFN & \cmark & & & \hres{75.2}{6.1} \\
XGBoost & & \cmark & & \hres{52.7}{10.3} \\
XGBoost & & & \cmark & \hres{44.9}{8.5} \\
TabPFN & \cmark & \cmark & & \textbf{\hres{86.8}{6.6}} \\
TabPFN & \cmark & & \cmark & \hres{81.4}{6.3} \\
TabPFN & \cmark & \cmark & \cmark & \textbf{\hres{86.3}{6.2}} \\
\bottomrule
\end{NiceTabular}
\end{table}

We tested tabular models on combinations of global features, reporting the results in \cref{tab:biomarkers_ablation_study}. Cardiac biomarkers are more predictive than vascular biomarkers ($+6.6$ on average for rows 2-3 and 4-5). Still, neither are informative on their own ($\le52.7$), and they benefit from taking into account medical records ($+35$ on average for rows 4-2 and 5-3). Surprisingly, adding vascular biomarkers to cardiac biomarkers and medical records does not significantly change results ($-0.5$ for rows 6-4). This suggests that discriminative information for PE risk stratification might be lost when summarising vascular graphs to a few biomarkers.

\smallskip
\noindent\textbf{Can GNNs leverage rich vascular graphs?}
To explore the hypothesis that vascular biomarkers discard important details from vascular graphs, we evaluated GNNs directly on the graphs, with and without global features. \Cref{tab:gnn_benchmark} benchmarks them against the best tabular model. GNNs without global features improve over tabular models on vascular biomarkers ($+8$ on average for rows 2 \& 6 vs.\ row 3 in \cref{tab:biomarkers_ablation_study}). Still, global features are necessary to reach competitive results. Compared to the impact of global features, differences between fusion methods across MPNN/GT are minor (test means all between $80.8$ and $83.8$). Still, the top two methods end up within statistical significance of TabPFN, even if they do not outperform it.

\begin{table}[tb]
\centering
\caption{Benchmark of GNNs, including models integrating global features, compared to tabular baseline. Rows without \textit{fusion method} indicate monomodal models.}
\label{tab:gnn_benchmark}
\begin{NiceTabular}{cccc cc cc}
\toprule
\Block{1-4}{Model}&&& & \Block{2-1}{Global \\ features} & \Block{2-1}{Vascular \\ graph} & \Block{1-2}{F1 (\%) $\uparrow$} \\
\cmidrule(lr){1-4} \cmidrule(lr){7-8}
Type & \makecell{Fusion \\ method} & \makecell{Best \\ backbone} & \makecell{\# \\ params} && & Validation & Test \\
\midrule
Tabular & -- & TabPFN & 7.2M & \cmark && -- & \textbf{\hres{86.3}{6.2}} \\
\midrule
\Block{4-1}{\rotate MPNN}
    & -- & GCN+VN & 22.3K && \cmark & \hres{65.5}{9.0} & \hres{55.9}{8.3} \\
    \cmidrule{2-8}
    & Early fusion & GCN+VN & 4.3K & \cmark & \cmark & \hres{88.1}{4.4} & \textbf{\hres{83.4}{4.4}} \\
    & Late fusion & GIN+VN & 89K & \cmark & \cmark & \hres{88.1}{4.6} & \hres{83.3}{5.1} \\
    & Virtual node & GIN+VN & 8.4K & \cmark & \cmark & \hres{87.6}{3.5} & \hres{81.4}{7.6} \\
\midrule
\Block{4-1}{\rotate GT}
    & -- & GPS & 23K && \cmark & \hres{65.6}{7.2} & \hres{50.2}{8.6} \\
    \cmidrule{2-8}
    & Early fusion & GPS & 35.3K & \cmark & \cmark & \hres{88.3}{4.6} & \hres{81.3}{4.8} \\
    & Late fusion & GPS & 97.4K & \cmark & \cmark & \hres{87.5}{5.3} & \hres{80.8}{6.4} \\
    & Cross-attention & GPS & 932K & \cmark & \cmark & \hres{89.2}{5.4} & \textbf{\hres{83.8}{4.9}} \\
\bottomrule
\end{NiceTabular}
\end{table}

\smallskip
\noindent\textbf{What explains suboptimal results with vascular graphs?}
As both vascular biomarkers and learning directly on vascular graphs counter-intuitively fail to improve risk predictions, we explore three limitations (overfitting, failure to learn, and biased labels) that could explain suboptimal performance.

First of all, given our sample size (n=353), GNNs might be overfitting, even if some are as small as 4.3K parameters. \Cref{tab:gnn_benchmark} does show a generalisation gap between validation and test of $-5.8$ on average for multimodal GNNs. However, this is in line with generalisation gaps reported on much larger datasets: $0.319/0.387$ $(+21\%)$ train/test MAE on ZINC with 12K graphs~\cite{dwivedi_benchmarking_2023}, and $84.79/77.07$ $(-7.72)$ val/test AUROC on \texttt{ogbg-molhiv} with 41K+ graphs~\cite{hu_open_2020}. Thus, GNNs do not overfit our dataset more than datasets two orders of magnitude larger.

Alternatively, the tree structure of vascular graphs may be fundamentally challenging for GNNs. We tested this by regressing vascular biomarkers more directly linked to local graph features than PE risk. \Cref{tab:gnn_sanity_check} compares TabPFN to GNNs for this, where \textit{Total Embolism Volume} and \textit{Qanadli score} are the simplest and most complex biomarkers to regress, respectively. Improvements over the tabular baseline ($+56$ on average for rows 2 \& 3 vs.\ 1) confirm that GNNs can learn patterns found only in vascular graphs and not in global features.

\begin{table}[tb]
\centering
\caption{Results for the regression of vascular biomarkers.}
\label{tab:gnn_sanity_check}
\begin{NiceTabular}{ccc ccc cc}
\toprule
\Block{1-3}{Model}&& & \Block{1-2}{Global features}& & \Block{2-1}{Vascular \\ graph} & \Block{1-2}{R\textsuperscript{2} (\%) $\uparrow$} \\
\cmidrule(lr){1-3} \cmidrule(lr){4-5} \cmidrule(lr){7-8}
\Block{1-2}{Type}& & \makecell{Best \\ backbone} & \makecell{Medical \\ records} & \makecell{Cardiac \\ biomarkers} & & \makecell{Total \\ Embolism \\ Volume} & \makecell{Qanadli \\ Score} \\
\midrule
\Block{1-1}{Baseline}
    & Tabular & TabPFN & \cmark & \cmark & & \hres{35.3}{20.3} & \hres{32.7}{5.6} \\
\midrule
\Block{2-1}{GNN}
    & MPNN & GCN+VN & && \cmark & \hres{95.7}{3.0} & \textbf{\hres{83.1}{4.2}} \\
    & GT & GPS & && \cmark & \textbf{\hres{98.2}{0.8}} & \textbf{\hres{82.0}{4.4}} \\
\bottomrule
\end{NiceTabular}
\end{table}

As for the labels, guideline risk groups might not exactly reflect real PE risk, and erroneous predictions from our models could be closer to real risk. This hypothesis is harder to test, as real PE risk in populations is typically quantified from 30-days mortality risk, too rare in our case for significant analysis (11 deaths for 353 patients). Because of this, we instead look at ambiguous cases, which we define as patients whose stratification changed between guidelines versions (2014 vs 2019), to check if they correlate with our models' errors. A large overlap could suggest that our models align with the guidelines overall, but disagree on difficult cases. However, 0 out of 64 errors from our best GNN (GPS with cross-attention) match the 39 ambiguous patients, refuting the initial hypothesis.

Considering the explanations ruled out above, the one that remains, although counter-intuitive, is that vascular graphs do not provide additional information beyond medical records and cardiac biomarkers to discriminate PE risk.

\section{Limitations \& Conclusions}
\label{sec:discussion+conclusion}

\noindent\textbf{Limitations. }
The focus on one dataset could be criticised; public datasets like PARSE (n=203)~\cite{luo_efficient_2024} and AirRC (n=254)~\cite{liu_custom_2025} also provide pulmonary vessel annotations but lack thrombi annotations, medical records and disease information. RSNA (n=12\,195)~\cite{colak_rsna_2021} targets PE and covers more cases, but lacks segmentations and complete medical records. Compared to these, our dataset provides a unique combination of vessel and thrombi segmentations and medical records, at a larger scale (n=353) than pixel-level annotated public datasets.

\smallskip
\noindent\textbf{Conclusions. }
This work presented the first study of the relevance of vascular graphs for pulmonary embolism risk stratification, enabled by an automatic CTPA processing pipeline. We extensively benchmarked methods to combine rich vascular graphs and medical records, from tabular models on global features to multimodal GNNs. The lack of improvements from vascular graphs lead us to investigate and rule out models and data limitations. Instead, our results refute clinical intuitions and suggest that vascular graphs do not contribute discriminative information for PE severity on top of simpler clinical biomarkers.

\begin{credits}
\subsubsection{\ackname} This work was supported by the French National Research Agency through the PERSEVERE (ANR-22-CE45-0018) and HistoGraph (ANR-23-CE45-0038) projects, and the LABEX PRIMES (ANR-11-LABX-0063) of Université de Lyon, within the program ``Investissements d'Avenir'' (ANR-11-IDEX-0007).

\subsubsection{\discintname} The authors have no relevant competing interests to declare.

\end{credits}

%
%
%
\bibliographystyle{splncs04}
\bibliography{references}

@misc{des-ligneris_patient-specific_2026,
    title = {A {Patient}-{Specific} {Pulmonary} {Arterial} {Tree} {Digital} {Twin} to {Extract} {Pulmonary} {Embolism} {Biomarkers}},
    url = {https://arxiv.org/abs/2605.28217v1},
    _doi = {10.48550/arXiv.2605.28217},
    author = {des Ligneris, Morgane and others},
    month = may,
	year = {2026},
    note = {{Under} review.}
}

@article{isensee_nnu-net_2021,
	title = {{nnU}-{Net}: a self-configuring method for deep learning-based biomedical image segmentation},
	volume = {18},
	shorttitle = {{nnU}-{Net}},
	doi = {10.1038/s41592-020-01008-z},
	number = {2},
	journal = {Nat Methods},
	author = {Isensee, Fabian and others},
	month = feb,
	year = {2021},
	pages = {203--211},
}

@inproceedings{gorishniy_revisiting_2021,
	title = {Revisiting {Deep} {Learning} {Models} for {Tabular} {Data}},
	booktitle = {Proc. {NeurIPS}},
	author = {Gorishniy, Yury and others},
	year = {2021},
	pages = {18932--18943},
}

@inproceedings{chen_xgboost_2016,
	title = {{XGBoost}: {A} {Scalable} {Tree} {Boosting} {System}},
	shorttitle = {{XGBoost}},
    doi = {10.1145/2939672.2939785},
	booktitle = {Proc. {KDD}},
	author = {Chen, Tianqi and Guestrin, Carlos},
	year = {2016},
	pages = {785--794},
}

@article{xu_multimodal_2023,
	title = {Multimodal {Learning} {With} {Transformers}: {A} {Survey}},
	volume = {45},
	shorttitle = {Multimodal {Learning} {With} {Transformers}},
    doi = {10.1109/TPAMI.2023.3275156},
	number = {10},
	journal = {IEEE Trans Pattern Anal Mach Intell},
	author = {Xu, Peng and Zhu, Xiatian and Clifton, David A.},
	month = oct,
	year = {2023},
	pages = {12113--12132},
}

@article{konstantinides_2019_2020,
	_title = {2019 {ESC} {Guidelines} for the diagnosis and management of acute pulmonary embolism developed in collaboration with the {European} {Respiratory} {Society} ({ERS}): {The} {Task} {Force} for the diagnosis and management of acute pulmonary embolism of the {European} {Society} of {Cardiology} ({ESC})},
    title = {2019 {ESC} {Guidelines} for the diagnosis and management of acute pulmonary embolism developed in collaboration with the {European} {Respiratory} {Society} ({ERS})},
	volume = {41},
    doi = {10.1093/eurheartj/ehz405},
	number = {4},
	journal = {European Heart Journal},
	author = {Konstantinides, Stavros V and {ESC Scientific Document Group} and others},
	month = jan,
	year = {2020},
	pages = {543--603},
}

@inproceedings{xu_how_2018,
	title = {How {Powerful} are {Graph} {Neural} {Networks}?},
	url = {https://openreview.net/forum?id=ryGs6iA5Km},
	booktitle = {Proc. {ICLR}},
	author = {Xu, Keyulu and others},
	year = {2018},
}

@inproceedings{kipf_semi-supervised_2017,
	title = {Semi-{Supervised} {Classification} with {Graph} {Convolutional} {Networks}},
	url = {https://openreview.net/forum?id=SJU4ayYgl},
	booktitle = {Proc. {ICLR}},
	author = {Kipf, Thomas N. and Welling, Max},
	year = {2017},
}

@article{ahmedt-aristizabal_graph-based_2021,
	title = {Graph-{Based} {Deep} {Learning} for {Medical} {Diagnosis} and {Analysis}: {Past}, {Present} and {Future}},
	volume = {21},
	shorttitle = {Graph-{Based} {Deep} {Learning} for {Medical} {Diagnosis} and {Analysis}},
	doi = {10.3390/s21144758},
	number = {14},
	journal = {Sensors},
	author = {Ahmedt-Aristizabal, David and others},
	month = jan,
	year = {2021},
	pages = {4758},
}

@inproceedings{brody_how_2021,
	title = {How {Attentive} are {Graph} {Attention} {Networks}?},
	url = {https://openreview.net/forum?id=F72ximsx7C1},
	booktitle = {Proc. {ICLR}},
	author = {Brody, Shaked and Alon, Uri and Yahav, Eran},
	year = {2021},
}

@inproceedings{rampasek_recipe_2022,
	title = {Recipe for a {General}, {Powerful}, {Scalable} {Graph} {Transformer}},
	booktitle = {Proc. {NeurIPS}},
	author = {Rampášek, Ladislav and others},
	year = {2022},
	pages = {14501--14515},
}

@article{khan_comprehensive_2025,
	title = {A {Comprehensive} {Survey} of {Foundation} {Models} in {Medicine}},
	doi = {10.1109/RBME.2025.3531360},
	journal = {IEEE Rev Biomed Eng},
	author = {Khan, Wasif and others},
	year = {2025},
	pages = {1--20},
}

@article{mastora_severity_2003,
	title = {Severity of acute pulmonary embolism: evaluation of a new spiral {CT} angiographic score in correlation with echocardiographic data},
	volume = {13},
	shorttitle = {Severity of acute pulmonary embolism},
	doi = {10.1007/s00330-002-1515-y},
	number = {1},
	journal = {Eur Radiol},
	author = {Mastora, Ioana and others},
	month = jan,
	year = {2003},
	pages = {29--35},
}

@article{qanadli_new_2001,
	title = {New {CT} {Index} to {Quantify} {Arterial} {Obstruction} in {Pulmonary}  {Embolism}},
	volume = {176},
	doi = {10.2214/ajr.176.6.1761415},
	number = {6},
	urldate = {2025-05-27},
	journal = {American Journal of Roentgenology},
	author = {Qanadli, Salah D. and others},
	month = jun,
	year = {2001},
	pages = {1415--1420},
}

@misc{pham_graph_2017,
	title = {Graph {Classification} via {Deep} {Learning} with {Virtual} {Nodes}},
	doi = {10.48550/arXiv.1708.04357},
	author = {Pham, Trang and others},
	month = aug,
	year = {2017},
}

@article{hollmann_accurate_2025,
	title = {Accurate predictions on small data with a tabular foundation model},
	volume = {637},
	doi = {10.1038/s41586-024-08328-6},
	number = {8045},
	journal = {Nature},
	author = {Hollmann, Noah and others},
	month = jan,
	year = {2025},
	pages = {319--326},
}

@inproceedings{fey_pyg_2025,
	title = {{PyG} 2.0: {Scalable} {Learning} on {Real} {World} {Graphs}},
	shorttitle = {{PyG} 2.0},
	url = {https://openreview.net/forum?id=DHHLkQvWqs},
    booktitle = {Proc. {TGL} {Workshop} @ {KDD}},
	author = {Fey, Matthias and others},
	year = {2025},
}

@article{liu_graph_2025,
	title = {Graph {Foundation} {Models}: {Concepts}, {Opportunities} and {Challenges}},
	volume = {47},
	shorttitle = {Graph {Foundation} {Models}},
	doi = {10.1109/TPAMI.2025.3548729},
	number = {6},
	journal = {IEEE Trans Pattern Anal Mach Intell},
	author = {Liu, Jiawei and others},
	month = jun,
	year = {2025},
	pages = {5023--5044},
}

@article{dwivedi_benchmarking_2023,
	title = {Benchmarking {Graph} {Neural} {Networks}},
	volume = {24},
	_url = {http://jmlr.org/papers/v24/22-0567.html},
	abstract = {In the last few years, graph neural networks (GNNs) have become the standard toolkit for analyzing and learning from data on graphs. This emerging field has witnessed an extensive growth of promising techniques that have been applied with success to computer science, mathematics, biology, physics and chemistry. But for any successful field to become mainstream and reliable, benchmarks must be developed to quantify progress. This led us in March 2020 to release a benchmark framework that i) comprises of a diverse collection of mathematical and real-world graphs, ii) enables fair model comparison with the same parameter budget to identify key architectures, iii) has an open-source, easy-to use and reproducible code infrastructure, and iv) is flexible for researchers to experiment with new theoretical ideas. As of December 2022, the GitHub repository has reached 2,000 stars and 380 forks, which demonstrates the utility of the proposed open-source framework through the wide usage by the GNN community. In this paper, we present an updated version of our benchmark with a concise presentation of the aforementioned framework characteristics, an additional medium-sized molecular dataset AQSOL, similar to the popular ZINC, but with a real-world measured chemical target, and discuss how this framework can be leveraged to explore new GNN designs and insights. As a proof of value of our benchmark, we study the case of graph positional encoding (PE) in GNNs, which was introduced with this benchmark and has since spurred interest of exploring more powerful PE for Transformers and GNNs in a robust experimental setting.},
	number = {43},
	journal = {Journal of Machine Learning Research},
	author = {Dwivedi, Vijay Prakash and others},
	year = {2023},
	pages = {1--48},
}

@inproceedings{brasoveanu_extending_2023,
	title = {Extending {Graph} {Neural} {Networks} with {Global} {Features}},
	url = {https://openreview.net/forum?id=aisVQy6R2k},
    booktitle = {{Learning} on {Graphs} ({LoG})},
	author = {Brasoveanu, Andrei Dragos and others},
	year = {2023},
}

@article{lutchyn_efficient_2025,
	title = {Efficient {Learning} of {Molecular} {Properties} {Using} {Graph} {Neural} {Networks} {Enhanced} with {Chemistry} {Knowledge}},
	doi = {10.1021/acsomega.5c07178},
	journal = {ACS Omega},
	author = {Lutchyn, Tetiana and Mardal, Marie and Ricaud, Benjamin},
	month = nov,
	year = {2025},
}

@inproceedings{akiba_optuna_2019,
	title = {Optuna: {A} {Next}-generation {Hyperparameter} {Optimization} {Framework}},
	shorttitle = {Optuna},
	doi = {10.1145/3292500.3330701},
	booktitle = {Proc. {KDD}},
	author = {Akiba, Takuya and others},
	year = {2019},
	pages = {2623--2631},
}

@inproceedings{sun_graph_2024,
	title = {Graph {Neural} {Network} based {Future} {Clinical} {Events} {Prediction} from {Invasive} {Coronary} {Angiography}},
	doi = {10.1109/ISBI56570.2024.10635813},
	booktitle = {Proc. {ISBI}},
	author = {Sun, Xiaowu and others},
	year = {2024},
	pages = {1--5},
}

@inproceedings{stym-popper_dafted_2025,
	title = {{DAFTED}: {Decoupled} {Asymmetric} {Fusion} of {Tabular} and {Echocardiographic} {Data} for {Cardiac} {Hypertension} {Diagnosis}},
	shorttitle = {{DAFTED}},
	url = {https://openreview.net/forum?id=ghhGImwv07},
	booktitle = {Proc. {MIDL}},
	author = {Stym-Popper, Jérémie and others},
	year = {2025},
}

@inproceedings{hu_open_2020,
	title = {Open {Graph} {Benchmark}: {Datasets} for {Machine} {Learning} on {Graphs}},
	shorttitle = {Open {Graph} {Benchmark}},
	booktitle = {Proc. {NeurIPS}},
	author = {Hu, Weihua and others},
	year = {2020},
	pages = {22118--22133},
}

@article{liu_custom_2025,
	title = {A {Custom} {Annotated} {Dataset} for {Segmentation} of {Pulmonary} {Veins}, {Arteries}, and {Airways}},
	volume = {12},
	doi = {10.1038/s41597-025-06074-6},
	number = {1},
	journal = {Sci Data},
	author = {Liu, Jian and others},
	month = nov,
	year = {2025},
	pages = {1806},
}

@misc{luo_efficient_2024,
	title = {Efficient automatic segmentation for multi-level pulmonary arteries: {The} {PARSE} challenge},
	shorttitle = {Efficient automatic segmentation for multi-level pulmonary arteries},
	doi = {10.48550/arXiv.2304.03708},
	author = {Luo, Gongning and others},
	month = aug,
	year = {2024},
}

@article{chu_deep_2025,
	title = {Deep learning-driven pulmonary artery and vein segmentation reveals demography-associated vasculature anatomical differences},
	volume = {16},
	doi = {10.1038/s41467-025-56505-6},
	number = {1},
	journal = {Nat Commun},
	publisher = {Nature Publishing Group},
	author = {Chu, Yuetan and others},
	month = mar,
	year = {2025},
	pages = {2262},
}

@article{colak_rsna_2021,
	title = {The {RSNA} {Pulmonary} {Embolism} {CT} {Dataset}},
	volume = {3},
	doi = {10.1148/ryai.2021200254},
	number = {2},
	journal = {Radiology: Artificial Intelligence},
	publisher = {Radiological Society of North America},
	author = {Colak, Errol et al., {For the RSNA-STR Annotators and Dataset Curation Contributors}},
	month = mar,
	year = {2021},
	pages = {e200254},
}

@article{djahnine_detection_2024,
	title = {Detection and severity quantification of pulmonary embolism with {3D} {CT} data using an automated deep learning-based artificial solution},
	volume = {105},
	doi = {10.1016/j.diii.2023.09.006},
	number = {3},
	journal = {Diagnostic and Interventional Imaging},
	author = {Djahnine, Aissam and others},
	month = mar,
	year = {2024},
	pages = {97--103},
}

@article{soffer_deep_2021,
	title = {Deep learning for pulmonary embolism detection on computed tomography pulmonary angiogram: a systematic review and meta-analysis},
	volume = {11},
	shorttitle = {Deep learning for pulmonary embolism detection on computed tomography pulmonary angiogram},
	doi = {10.1038/s41598-021-95249-3},
	number = {1},
	journal = {Sci Rep},
	author = {Soffer, Shelly and others},
	month = aug,
	year = {2021},
	pages = {15814},
}

@article{wasserthal_totalsegmentator_2023,
	title = {{TotalSegmentator}: {Robust} {Segmentation} of 104 {Anatomic} {Structures} in {CT} {Images}},
	volume = {5},
	shorttitle = {{TotalSegmentator}},
	doi = {10.1148/ryai.230024},
	number = {5},
	journal = {Radiology: Artificial Intelligence},
	author = {Wasserthal, Jakob and others},
	month = sep,
	year = {2023},
	pages = {e230024},
}

@article{guarrasi_systematic_2025,
	title = {A systematic review of intermediate fusion in multimodal deep learning for biomedical applications},
	volume = {158},
	doi = {10.1016/j.imavis.2025.105509},
	journal = {Image Vis Comput},
	author = {Guarrasi, Valerio and others},
	month = may,
	year = {2025},
	pages = {105509},
}

@article{jimenez_simplification_2010,
	title = {Simplification of the {Pulmonary} {Embolism} {Severity} {Index} for {Prognostication} in {Patients} {With} {Acute} {Symptomatic} {Pulmonary} {Embolism}},
	volume = {170},
	doi = {10.1001/archinternmed.2010.199},
	number = {15},
	journal = {Arch Intern Med},
	author = {Jiménez, David and {RIETE Investigators} and others},
	month = aug,
	year = {2010},
	pages = {1383--1389},
}

@article{liu_joint_2021,
	title = {Joint analysis of {D}-dimer, {N}-terminal pro b-type natriuretic peptide, and cardiac troponin {I} on predicting acute pulmonary embolism relapse and mortality},
	volume = {11},
	doi = {10.1038/s41598-021-94346-7},
	number = {1},
	journal = {Sci Rep},
	author = {Liu, Xiaoyu and others},
	month = jul,
	year = {2021},
	pages = {14909},
}
\end{document}